%% file: main.tex
\documentclass[10pt,twocolumn,letterpaper]{article}

\usepackage{cvpr}              % To produce the CAMERA-READY version
\input{preamble}

\definecolor{cvprblue}{rgb}{0.21,0.49,0.74}
\usepackage[pagebackref,breaklinks,colorlinks,allcolors=cvprblue]{hyperref}

\def\paperID{6} % *** Enter the Paper ID here
\def\confName{CVPR}
\def\confYear{2026}

\title{Fine-Tuning Large Language Models for Cooperative Tactical Deconfliction of Small Unmanned Aerial Systems}

\author{
Iman Sharifi$^{*}$ \qquad Alex Zongo$^{*}$ \qquad Peng Wei\\
George Washington University\\
{\ttfamily \small \{i.sharifi,a.zongo,pwei\}@gwu.edu}\\
{\small $^{*}$Equal contribution}
}

\begin{document}
\maketitle
\input{sec/0_abstract}    
\input{sec/1_Introduction}
\input{sec/2_related_works}
\input{sec/3_Problem_formulation_method}

\input{sec/4_data_gen_pipeline}

\input{sec/5_LLM_selection_finetuning}

\input{sec/6_Experiments}
\input{sec/7_conclusion}
{
    \small
    \bibliographystyle{ieeenat_fullname}
    \bibliography{main}
}
\input{sec/Supplementary_Material}

\end{document}

%% file: preamble.tex
\usepackage{booktabs}
\usepackage{tabularx}
\usepackage{multicol}
\usepackage{multirow}
\usepackage{makecell}
\usepackage{listings}
\usepackage{xcolor}

\lstdefinestyle{stateSnapshotCVPR}{
  basicstyle=\ttfamily\scriptsize,
  columns=fullflexible,
  keepspaces=true,
  showstringspaces=false,
  breaklines=true,
  breakatwhitespace=true,     % wrap at spaces when possible
  breakautoindent=true,
  postbreak=\mbox{\textcolor{gray}{$\hookrightarrow$}\space},
  frame=single,
  framerule=0.3pt,
  rulecolor=\color{black!20},
  xleftmargin=0.6em,
  xrightmargin=0.2em,
  aboveskip=0.4em,
  belowskip=0.2em,
  captionpos=b,
  lineskip=0pt,
  literate={,}{{,}}1 {:}{{:\hspace{0pt}}}1 {_}{{\_\hspace{0pt}}}1
           {(}{{(\hspace{0pt}}}1 {)}{{)\hspace{0pt}}}1
           {-}{{-\hspace{0pt}}}1 {.}{{.\hspace{0pt}}}1
}

\makeatletter
\renewcommand{\maketitlesupplementary}{%
  \par\begingroup
  \renewcommand{\thefootnote}{\fnsymbol{footnote}}%
  \def\@makefnmark{\hbox{\@textsuperscript{\@thefnmark}}}%
  \long\def\@makefntext##1{\parindent 1em\noindent
    \hb@xt@1.8em{\hss\@textsuperscript{\@thefnmark}}##1}%
  \thispagestyle{empty}%
  \begin{center}
    {\Large\bfseries \@title \par}
    \vskip 0.8em
    {\large Supplementary Material\par}
    \vskip 0.8em
    {\normalsize \@author \par}
  \end{center}
  \vskip 1em
  \setcounter{footnote}{0}%
  \renewcommand{\thefootnote}{\arabic{footnote}}%
  \endgroup
}
\makeatother

%% file: sec/0_abstract.tex
\begin{abstract}
The growing deployment of small Unmanned Aerial Systems (sUASs) in low-altitude airspaces has increased the need for reliable tactical deconfliction under safety-critical constraints. Tactical deconfliction involves short-horizon decision-making in dense, partially observable, and heterogeneous multi-agent environments, where both cooperative separation assurance and operational efficiency must be maintained. While Large Language Models (LLMs) exhibit strong reasoning capabilities, their direct application to air traffic control remains limited by insufficient domain grounding and unpredictable output inconsistency. This paper investigates LLMs as decision-makers in cooperative multi-agent tactical deconfliction using fine-tuning strategies that align model outputs to human operator heuristics. We propose a simulation-to-language data generation pipeline based on the BlueSky air traffic simulator that produces rule-consistent deconfliction datasets reflecting established safety practices. A pretrained Qwen-Math-7B model is fine-tuned using two parameter-efficient strategies: supervised fine-tuning with Low-Rank Adaptation (LoRA) and preference-based fine-tuning combining LoRA with Group-Relative Policy Optimization (GRPO). Experimental results on validation datasets and closed-loop simulations demonstrate that supervised LoRA fine-tuning substantially improves decision accuracy, consistency, and separation performance compared to the pretrained LLM, with significant reductions in near mid-air collisions. GRPO provides additional coordination benefits but exhibits reduced robustness when interacting with heterogeneous agent \mbox{policies}.
\end{abstract}

%% file: sec/1_Introduction.tex
\section{Introduction}
\label{sec:intro}
The rapid growth in civil and commercial deployment of small Unmanned Aerial Systems (sUASs), including package delivery, infrastructure inspection, and emergency response, has intensified the demand for safe and efficient operations in low-altitude, shared airspaces~\cite{faa_dallas_history_2024,marquand_bvlos_dallas_2024}. As traffic \mbox{density} increases, conflicts between vehicles become inevitable, particularly near intersections, merging corridors, and other constrained airspace regions. Tactical deconfliction, which involves real-time, short-horizon decision-making that maintains safe separation while preserving operational efficiency, has therefore emerged as a central challenge in UAS traffic management (UTM) ecosystems. Unlike strategic planning~\cite{liu2023strategic} or trajectory optimization, tactical deconfliction must operate under strict time constraints, partial observability, and complex multi-agent interactions, where delayed or overly conservative decisions can significantly degrade both safety and traffic throughput~\cite{ribeiro2020review}. Rule-based approaches lack flexibility and scalability~\cite{HOEKSTRA2002215,ssd}, while optimization-based and learning-based methods often struggle with latency, \mbox{robustness}, or interpretability under safety-critical constraints~\cite{optimization_based_tactical_deconfliction,AI_uav_deconfliction_survey}.

% Recent advances in Large Language Models (LLMs) have demonstrated strong capabilities in logical reasoning~\cite{llm_logical_reasoning,imani2023mathpromptermathematicalreasoningusing}, contextual understanding~\cite{llm_zero_shot_reasoners}, and sequential decision-making across a wide range of domains~\cite{wang2024voyager}. In addition, LLMs have shown an ability to emulate human-like \mbox{behaviors} and decision patterns in complex environments~\cite{lin2024learningmodelworldlanguage}. These properties make LLMs a promising approach for tactical deconfliction, which fundamentally requires interpreting dynamic multi-agent states, prioritizing safety constraints, and selecting context-aware actions under uncertainty~\cite{cheng2025language}. However, general-purpose LLMs are not inherently suited for safety-critical aviation tasks~\cite{yuan2025nextgenerationllmuavnatural}. When applied in a zero-shot or prompt-engineered manner, their outputs can be inconsistent, highly sensitive to prompt structure~\cite{guan2025ordereffectinvestigatingprompt, errica-etal-2025-wrong}, and misaligned with established human safety norms~\cite{jiang2025training}. Furthermore, LLMs trained primarily on open-domain text lack exposure to domain-specific operational trade-offs~\cite{jiang2025training}, such as balancing separation assurance against mission efficiency in dense airspaces. These limitations motivate the need for systematic alignment between LLM decision-making behavior and human tactical reasoning in sUAS operations.

Recent advances in Large Language Models (LLMs) have shown strong capabilities in reasoning~\cite{llm_logical_reasoning,imani2023mathpromptermathematicalreasoningusing}, contextual understanding~\cite{llm_zero_shot_reasoners}, and sequential decision-making~\cite{wang2024voyager}, making them a promising candidate for tactical deconfliction in dense, uncertain multi-agent airspaces~\cite{cheng2025language}. Yet, general-purpose LLMs are not designed for safety-critical aviation~\cite{yuan2025nextgenerationllmuavnatural}: zero-shot or prompt-based use can yield inconsistent and prompt-sensitive outputs~\cite{guan2025ordereffectinvestigatingprompt, errica-etal-2025-wrong}, misaligned with human safety norms~\cite{jiang2025training}, and uninformed about domain-specific trade-offs~\cite{jiang2025training}. These limitations motivate systematic alignment of LLM behavior with human tactical reasoning in sUAS operations.

% Human operators, such as air traffic controllers and experienced pilots, resolve conflicts by applying implicit safety principles: prioritizing separation, anticipating other agents' intent, and reasoning over short temporal horizons rather than optimizing explicit reward functions~\cite{loft2007modeling}. This observation motivates the use of human-aligned datasets, wherein expert knowledge is encoded as logical rules, as a foundation for adapting LLMs to tactical deconfliction tasks. By fine-tuning LLMs on data that reflects human judgments, preferences~\cite{ouyang2022training}, and safety-oriented decision patterns, we can imbue these models with domain-appropriate reasoning priors that are difficult to capture through handcrafted rules or reward shaping alone~\cite{amodei2016concreteproblemsaisafety}. In contrast to multi-agent reinforcement learning approaches that learn policies through trial-and-error interaction~\cite{8917217}, human-aligned fine-tuning emphasizes transferring human tactical judgment directly into LLM inference behavior while preserving interpretability and operational consistency.
%~\cite{pmlr-v205-ichter23a}.

Human experts (e.g., air traffic controllers and experienced pilots) resolve conflicts by applying implicit safety principles, i.e., prioritizing separation, anticipating others' intent and reasoning over short horizons, rather than optimizing explicit reward functions~\cite{loft2007modeling}. We therefore advocate leveraging human-aligned datasets that encode expert knowledge as logical rules, and fine-tuning LLMs to transfer these judgments and preferences into inference-time behavior~\cite{ouyang2022training}. Compared to trial-and-error multi-agent reinforcement learning~\cite{8917217}, human-aligned fine-tuning can inject domain-appropriate reasoning priors while promoting interpretability and behavioral consistency~\cite{amodei2016concreteproblemsaisafety}.

In this paper, we present a simulation-to-language dataset generation pipeline that enables systematic learning of human-aligned cooperative tactical deconfliction behaviors from high-fidelity air traffic simulations. The proposed pipeline generates diverse multi-agent scenarios, encodes human tactical knowledge through logical rules, and transforms raw simulation data into structured prompt--response pairs suitable for training LLMs. Using this dataset, we study two complementary fine-tuning strategies for adapting pre-trained LLMs to tactical deconfliction in multi-agent sUAS environments. To the best of our knowledge, this work constitutes the first systematic investigation of fine-tuned LLMs for tactical deconfliction evaluated both on held-out datasets and in closed-loop air traffic simulations. The main contributions of this work are as follows:
\begin{itemize}
    \item We develop a simulation-to-language dataset generation pipeline based on the BlueSky air traffic simulator~\cite{bluesky_paper} that enables rapid construction of large-scale, rule-consistent tactical deconfliction datasets, allowing LLMs to internalize human safety heuristics and operational preferences.
    \item We demonstrate that parameter-efficient Supervised Fine-Tuning (SFT) with Low-Rank Adaptation (LoRA)~\cite{hu2022lora} substantially improves LLM decision accuracy, behavioral consistency, and separation safety compared to a pretrained baseline, as validated through both offline evaluation and closed-loop simulation.
    \item We evaluate the performance of the preference-based fine-tuning using Group-Relative Policy Optimization (GRPO) compared to SFT, providing insight into the strengths and limitations of reinforcement-style alignment for tactical deconfliction.
\end{itemize}

The remainder of this paper is organized as follows. Section~\ref{sec:related_work} reviews the related works. Section~\ref{sec:problem_formulation_methodology} formulates the tactical deconfliction problem and the fine-tuning strategies. Section~\ref{sec:presimulation} describes the simulation-to-language dataset generator pipeline. Section~\ref{sec:llm_selection_finetuning} elaborates on the two fine-tuning strategies. Section~\ref{sec:results} reports experimental results and comparative evaluations. Finally, Section~\ref{sec:conclusion} draws conclusions.

%% file: sec/2_related_works.tex
\section{Related Work}
\label{sec:related_work}

Recent research has explored the application of LLMs to air traffic control. These efforts have primarily positioned LLMs as high-level reasoning, interface, or knowledge-support components rather than direct low-level controllers. Several \mbox{studies} employ LLMs as natural-language interfaces integrated with existing conflict resolution solvers, allowing air traffic controllers to express preferences and constraints while preserving safety guarantees through restricted LLM outputs limited to filtering or ranking candidate solutions~\cite{LIU2024100024}. Other work investigates LLMs as embodied or tool-augmented agents capable of directly issuing control commands in simulation environments~\cite{andriuvskevivcius2024automatic}, often augmented with role decomposition or experience libraries to improve reasoning consistency. Complementary efforts leverage LLMs for air traffic scenario generation~\cite{Gould2025AirTrafficGenCA}, aviation-domain knowledge modeling~\cite{Wang2023AviationGPTAL}, and systematic evaluation of LLM reliability, recall, and reasoning performance in aviation contexts~\cite{ge2025aviation}.  

These studies reveal recurring limitations, including sensitivity to prompt structure, hallucinations, limited recall, inference latency, and the absence of explicit alignment with human operational preferences, that pose significant challenges for real-time, safety-critical tactical deconfliction. Unlike prior LLM-based ATC approaches that rely on zero-shot prompting or prompt engineering with function-calling at inference time~\cite{andriuvskevivcius2024automatic}, this paper adopts a systematic fine-tuning strategy grounded in human-aligned data and achieves near-real-time performance. By adapting pre-trained LLMs through parameter-efficient fine-tuning and preference-aware optimization on rule-consistent, simulator-generated datasets, our approach positions LLMs as human-aligned tactical decision-makers rather than free-form reasoning agents. This directly addresses the reliability and consistency concerns highlighted in existing LLM-based ATC research.

%% file: sec/3_Problem_formulation_method.tex
\section{Problem Formulation and Methodology}
\label{sec:problem_formulation_methodology}

\subsection{Tactical Deconfliction with LLM-based Policies}
\label{sec:tactical_deconfliction_llm}

We consider a tactical deconfliction problem in a shared low-altitude airspace populated by multiple sUASs with heterogeneous configurations and decision-making policies. Agents may differ in kinematic limits, sensing capabilities, maneuverability, and onboard autonomy architecture. The objective of tactical deconfliction is to maintain, cooperatively, safe separation while minimizing unnecessary deviations from nominal mission trajectories. Decisions must be made in real time under partial observability and amid complex multi-agent interactions.

Rather than addressing deconfliction through continuous control or trajectory optimization, we formulate the problem at the policy level. At each decision step, an agent observes a structured representation of the surrounding environment, including its own state, nearby traffic information, and safety constraints. Based on this context, the agent selects a discrete tactical action, such as accelerating, maintaining speed, or decelerating. The LLM serves as a high-level policy that maps structured agent state descriptions to tactical decisions. The LLM outputs abstract actions that are subsequently executed by UAS flight control modules, allowing the model to reason over heterogeneous agent interactions and implicit safety priorities without requiring access to explicit models of low-level dynamics. To align the LLM behavior with domain-specific operational requirements, we fine-tuned the model on a large-scale dataset spanning diverse traffic scenarios.
% To improve the overall efficiency of the LLMs, they are fine-tuned on large datasets, including several air traffic scenarios.

\begin{figure*}
    \centering
    \includegraphics[width=\linewidth, trim={6mm 0mm 6mm 0mm}, clip]{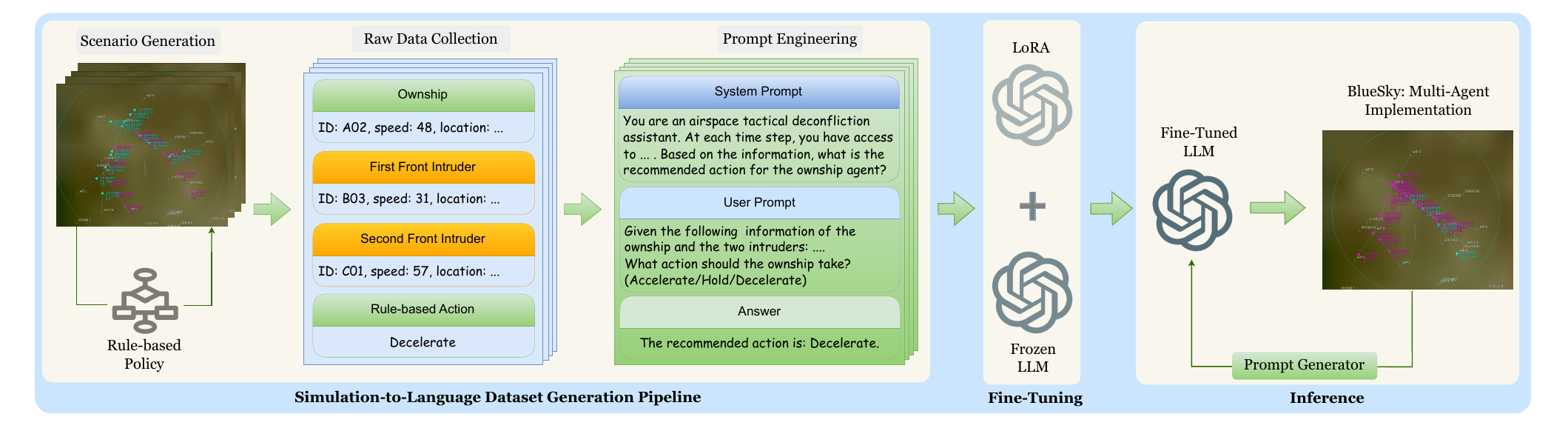}
    \caption{\textbf{Architecture overview.} The figure illustrates the end-to-end system architecture and the role of the proposed simulation-to-language dataset generation pipeline. Multi-agent traffic scenarios are generated in the BlueSky simulator, from which raw state data are extracted and converted into structured natural-language prompts using rule-based supervision. The resulting prompt--response pairs constitute the training dataset for LoRA-based fine-tuning. At deployment, the fine-tuned LLM generates tactical actions for multiple agents, which are executed in BlueSky, closing the simulation loop.}

    \label{fig:llm_atc}
\end{figure*}

\subsection{Fine-Tuning Strategies}
\label{sec:finetuning_strategies}

\subsubsection{Supervised Fine-Tuning (SFT)}
\label{sec:sft_lora}

This first strategy adapts a pre-trained LLM to tactical deconfliction through supervised learning on human-aligned, rule-consistent datasets. Each training sample consists of a structured description of the ownship’s local traffic context paired with a target tactical action derived from human-designed safety rules. Given a dataset $\mathcal{D} = \{(x_i, y_i)\}_{i=1}^{N}$, where $x_i$ denotes the ownship context and $y_i$ the corresponding target action, the objective is to maximize the conditional likelihood of human-aligned decisions under the fine-tuned model.

Formally, SFT minimizes the negative log-likelihood loss
\(\mathcal{L}_{\text{SFT}} = - \mathbb{E}_{(x,y) \sim \mathcal{D}} \left[ \log p_\theta(y \mid x) \right],\)
where $p_\theta$ denotes the LLM parameterized by $\theta$. Minimizing this loss transfers human decision heuristics into the model's inference behavior, encouraging consistent reproduction of safety-oriented tactical actions across similar agent state configurations. 
% Minimizing this loss encourages the model to consistently reproduce safety-oriented tactical actions for similar agent state configurations, effectively transferring human decision heuristics into the model’s inference behavior.

To enable efficient domain adaptation without updating the full parameter set of the LLM, we employ Low-Rank Adaptation (LoRA)~\cite{hu2022lora}, as shown in Figure~\ref{fig:llm_atc}, which injects trainable low-rank updates into selected projection layers while keeping the pretrained weights frozen. This parameter-efficient design enables scalable adaptation while preserving the general reasoning capabilities of the base model.

\subsubsection{Group-Relative Policy Optimization (GRPO)}
\label{sec:grpo_lora}

The second fine-tuning strategy employs GRPO, a preference-based alignment method that refines LLM behavior using sampled candidate actions and scalar reward feedback. For a given agent context $x$, the pretrained LLM generates a set of candidate tactical responses $\{y^{(1)}, \ldots, y^{(K)}\}$ via high-temperature sampling, promoting decision exploration beyond deterministic imitation.

Each candidate response is evaluated using a task-specific reward function $R(x, y)$ that encodes human-aligned safety rules and operational preferences, assigning higher scores to actions that maintain separation, respect right-of-way precedence, and favor conservative, interpretable maneuvers. These rewards are used to compute a group-relative advantage
\begin{equation}
\hat{A}^{(k)} = R(x, y^{(k)}) - \frac{1}{K} \sum_{j=1}^{K} R(x, y^{(j)}),
\label{eq:advantage}
\end{equation}
which measures the relative quality of each response within the sampled group.

Model parameters are then updated using an objective based on Proximal Policy Optimization (PPO) loss that increases the likelihood of higher-advantage responses while constraining policy updates for stability. The resulting GRPO loss is given by
\begin{equation} \begin{aligned} \mathcal{L}_{\text{GRPO}} = - \mathbb{E}_{x, y^{(k)}} [ & \min ( \rho^{(k)} \hat{A}^{(k)}, \\ & \text{clip}(\rho^{(k)}, 1 - \epsilon, 1 + \epsilon)\hat{A}^{(k)} ) ], \end{aligned} \end{equation}
where $\rho^{(k)} = \frac{p_\theta(y^{(k)} \mid x)}{p_{\theta_{\text{old}}}(y^{(k)} \mid x)}$ denotes the likelihood ratio between the updated and previous policies, and $\epsilon$ is a clipping parameter. As in SFT, GRPO updates are applied exclusively through LoRA parameters, leaving the base model unchanged.

By combining stochastic exploration, rule-based reward evaluation, and PPO-style optimization, GRPO enables preference-driven refinement of LLM decision-making beyond direct imitation. Unlike SFT, which enforces alignment through supervised reproduction of human actions, GRPO encourages relative improvement among competing candidate responses. This distinction enables a principled comparison between imitation-based and preference-based alignment for safety-critical tactical deconfliction.

%% file: sec/4_data_gen_pipeline.tex
\section{Dataset Generation Pipeline}
\label{sec:presimulation}

As major companies increasingly deploy sUAS fleets in shared airspace, safety- and privacy-related constraints have become central considerations to their operational frameworks. Due to proprietary concerns and regulatory sensitivities, high-fidelity operational data relevant to tactical deconfliction is rarely publicized, limiting the availability of real-world datasets for learning-based methods. This lack of accessible data poses a fundamental barrier to the development and evaluation of data-driven deconfliction policies, which typically rely on large-scale, representative training corpora. To address this challenge, we design a simulation-based dataset generation pipeline that enables systematic, privacy-preserving collection of human-aligned tactical decision data, while remaining extensible to future integration with real-world observations as such data become available.

Thus, to collect trainable datasets including pairs of prompts and rule-based responses, we designed a simulation-to-language pipeline that generates scenarios and converts them to trainable prompt-answer pairs. Figure~\ref{fig:llm_atc} illustrates an overview of the pipeline, in which we initially collect a series of high-fidelity multi-agent simulations using the BlueSky Air Traffic Simulator~\cite{bluesky_paper}. The simulation environment was configured to emulate low-altitude airspace over the city of Frisco, Texas, a representative urban hub for drone delivery operations. The dataset generation pipeline includes the following stages: 

\textit{\textbf{Scenario Configurations:}} 
We generated diverse multi-agent flight scenarios to capture the traffic complexity and interaction patterns characteristic of urban low-altitude operations. Each scenario involves 20--30 sUASs operating concurrently in shared airspace and includes two merging points and one intersection, reflecting common bottlenecks in drone delivery corridors. To introduce variability in traffic density and agent state geometry, the number of active flight routes per scenario was randomly varied between four and six, producing heterogeneous traffic flows with intersecting and merging trajectories.

To model realistic fleet diversity, we defined two distinct agent configurations characterized by different speed limits, acceleration capabilities, and sensing ranges. These configurations represent heterogeneous vehicle capabilities commonly observed across different drone operators and enable systematic evaluation of an LLM’s ability to generalize across agents with varying dynamics. Specifically, we consider configurations \texttt{X} and \texttt{Y}, where configuration \texttt{X} exhibits stronger kinematic and sensing capabilities than configuration \texttt{Y}. The speed and acceleration limits for configurations \texttt{X} and \texttt{Y} are selected based on the performance specifications of the Google Wing Hummingbird drone~\cite{wing2024howItWorks} and the Amazon MK30 drone~\cite{faa2025amazonMK30}, respectively. Sensing ranges reflect current technological constraints associated with Remote ID-based communication or radar-based detection systems, ensuring realistic perception asymmetry among agents.

Table~\ref{tab:config-specs} summarizes the kinematic and sensing specifications for each configuration. By incorporating heterogeneous vehicle capabilities and structurally complex airspace layouts, the proposed scenario design yields a challenging and representative testbed for learning and evaluating cooperative tactical deconfliction policies under realistic drone delivery operations.

\begin{table}[!t]
\caption{Kinematic and sensing specifications for UAS configurations \texttt{X} (strong) and \texttt{Y} (weak).}

\label{tab:config-specs}
\centering
\scriptsize
\setlength{\tabcolsep}{6.5pt}
\renewcommand{\arraystretch}{1.3}
\begin{tabular}{lccc}
\toprule
\multirow{2}{*}{\textbf{Parameter}} & \multirow{2}{*}{\textbf{Notation}} & \multicolumn{2}{c}{\textbf{Configuration}} \\
\cmidrule{3-4} 
& & \textbf{\texttt{X} (strong)} & \textbf{\texttt{Y} (weak)} \\
\midrule
\makecell{Speed Range \\ (m/s)} & $[v_{\text{min}},v_{\text{max}}]$ & $[0, 44.88]$ & $[0, 30.12]$ \\
\makecell{Acceleration \\ (m/s$^2$)}  & $\Delta v/\Delta t$ & $\{-1.71, 0,  1.71\}$ & $\{-1.02, 0, 1.02\}$ \\
\makecell{Sensing Range \\ (m)}  & $\mathcal{R}$ & $1000$ & $750$ \\
\midrule
\bottomrule
\end{tabular}
\end{table}

\textit{\textbf{Rule-Based Policy Design:}} 
To generate human-aligned supervisory signals for tactical deconfliction, we designed a deterministic rule-based policy that enforces safe separation across all simulated scenarios. The policy is intended to emulate human pilot or controller reasoning by prescribing actions through interpretable if--then rules derived from operational heuristics.

At each decision step, the policy evaluates the local traffic context of a given agent (referred to as the ownship) and selects an appropriate tactical action based on multiple state-dependent factors. These include the ownship’s current and desired speeds, distance to the next waypoint, the number of nearby intruders, and relative spatial relationships with those intruders. To balance computational efficiency with behavioral fidelity, only the two closest front intruders are considered, as they typically represent the most critical conflict threats in dense airspace configurations. %Having conflicts with the ownship agent, front intruders are the agents closer to the next bottleneck waypoint.  

The policy further distinguishes between intruders operating on the same route and those on intersecting or merging routes. The policy is then enabled to modulate maneuver aggressiveness based on conflict geometry. Based on the evaluated conditions, the rule engine outputs one of three discrete tactical actions: \textit{Accelerate}, \textit{Hold}, or \textit{Decelerate}. These actions serve as human-aligned supervisory labels for dataset generation rather than as optimized control commands.

The complete rule hierarchy, decision thresholds, and tie-breaking logic are detailed in Supplementary Material (Appendix A).
% Appendix~\ref{app:rb_policy}.

\textit{\textbf{Raw Data Collection:}} 
For each simulation episode, state information was recorded for every ownship at discrete time steps. The collected data include the ownship’s position, velocity, heading, route identifier, and distance to the next waypoint, along with detailed information about the two closest front intruder agents, such as their relative positions, velocities, and distances to their respective waypoints.

To support flexible prompt construction and preserve contextual richness, both essential and supplementary attributes were retained during data logging. This design choice ensures that no potentially relevant information is lost during post-processing and allows multiple prompt formulations to be explored without re-running simulations. The resulting dataset captures dynamic multi-agent interactions across thousands of time steps and diverse traffic configurations.
An example of the raw observation record is provided in Supplementary Material (Appendix B). In total, over 38K state--action samples were collected in under 10 minutes. The data collection pipeline is fully modular, enabling additional scenarios and samples to be generated as needed.

\begin{figure*}[t!] 
\centering 
\begin{subfigure}[b]{0.45\textwidth}
    \includegraphics[width=\textwidth]{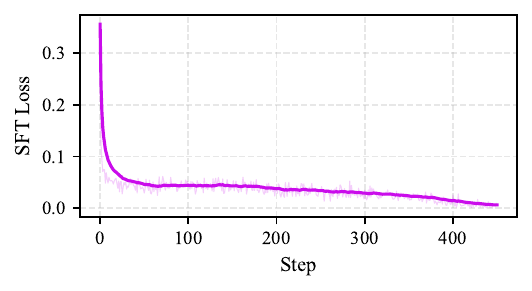}
    \caption{Loss curve during SFT training.}
    \label{fig:sft-curves}
\end{subfigure}
\hfill
\begin{subfigure}[b]{0.45\textwidth}
    \includegraphics[width=\textwidth]{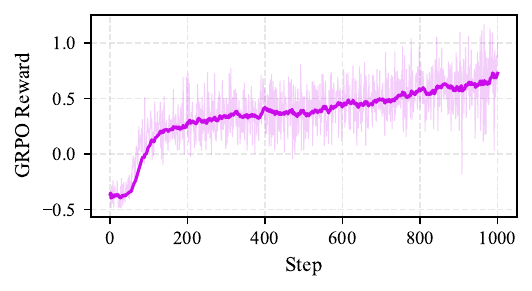}
    \caption{Reward progression during GRPO fine-tuning.}
    \label{fig:grpo-reward}
\end{subfigure}
\caption{Training effectiveness of fine-tuning methods. (a) shows the supervised learning progress through loss reduction, hence accuracy increase, while (b) shows the GRPO reward evolution across training iterations.}
\label{fig:training-curves}
\end{figure*}

\textit{\textbf{Prompt Engineering:}} 
As illustrated in Figure~\ref{fig:llm_atc}, following data collection, the raw numerical and categorical state information was transformed into structured natural-language prompts suitable for LLM training. Each prompt consists of two components: a \emph{system prompt}, which defines the model’s operational role and high-level objectives (e.g., ensuring safe separation in shared airspace), and a \emph{user prompt}, which describes the current local traffic situation of the ownship and nearby intruders in natural language. An illustrative example of the prompt format is presented in Supplementary Material (Appendix C).
% Appendix~\ref{app:prompt_example}.

This translation process converts low-level simulator states into human-readable descriptions that emphasize relative relationships, safety-relevant constraints, and decision context. As a result, the LLM is encouraged to infer tactical reasoning patterns rather than merely learning numerical correlations. The prompt format is kept consistent across training and inference to ensure behavioral stability.

The resulting pipeline produces a large-scale, context-rich dataset that embeds human tactical reasoning through interpretable rule-based supervision. The pipeline is computationally efficient, enabling rapid generation of training data and straightforward scaling to larger datasets as needed. Moreover, the pipeline's modular architecture allows both the rule-based policy and prompt engineering strategy to be replaced without modifying the underlying simulation infrastructure. By grounding LLM training data in high-fidelity simulations while maintaining flexibility and scalability, the pipeline provides a principled and extensible foundation for aligning LLM inference behavior with safety-critical deconfliction objectives.

\begin{figure*}[t]
    \centering
    \begin{subfigure}[t]{0.32\textwidth}
        \centering
        \includegraphics[width=\linewidth]{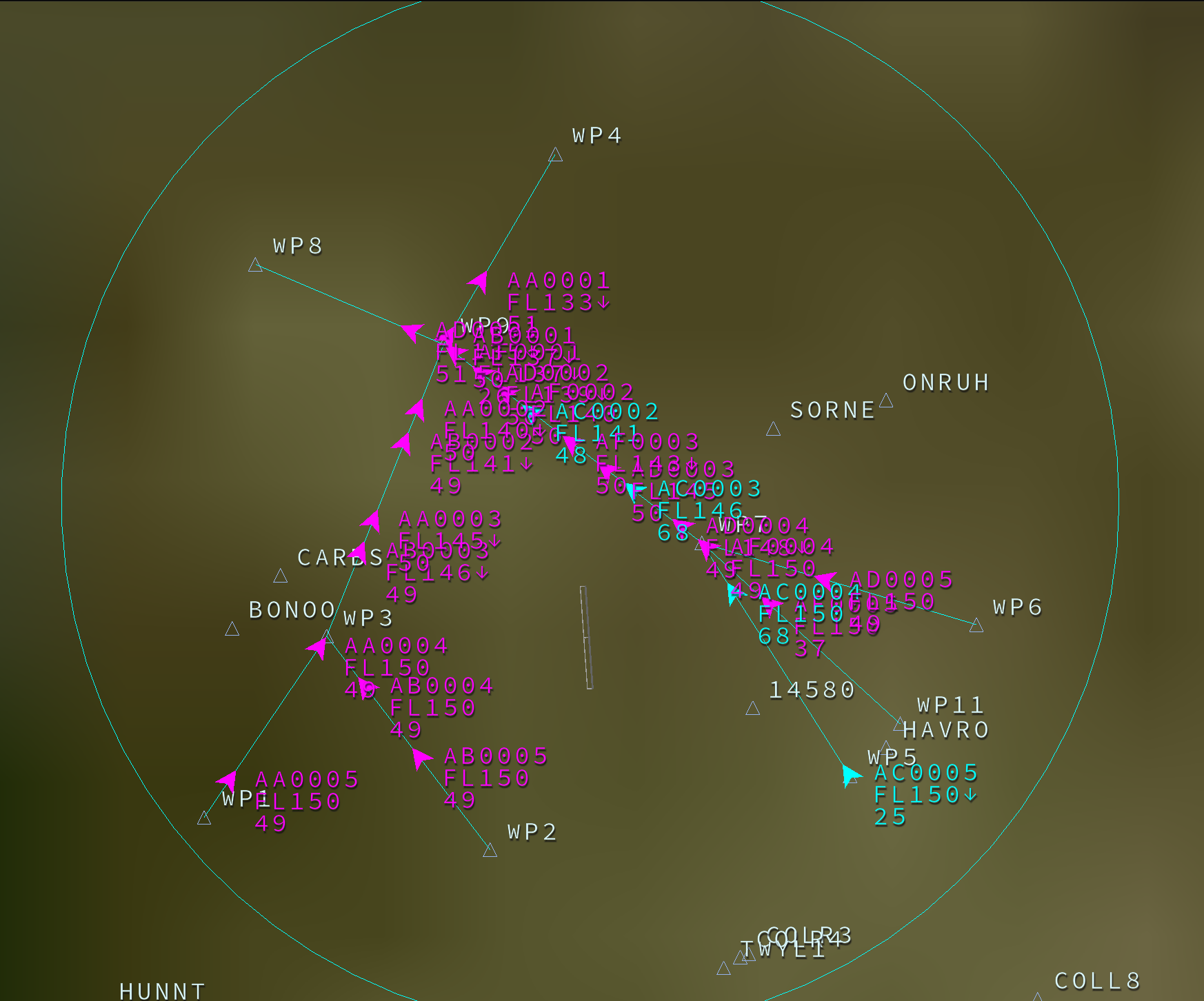}
        \caption{Scenario A}
        \label{fig:scenA}
    \end{subfigure}\hfill
    \begin{subfigure}[t]{0.32\textwidth}
        \centering
        \includegraphics[width=\linewidth]{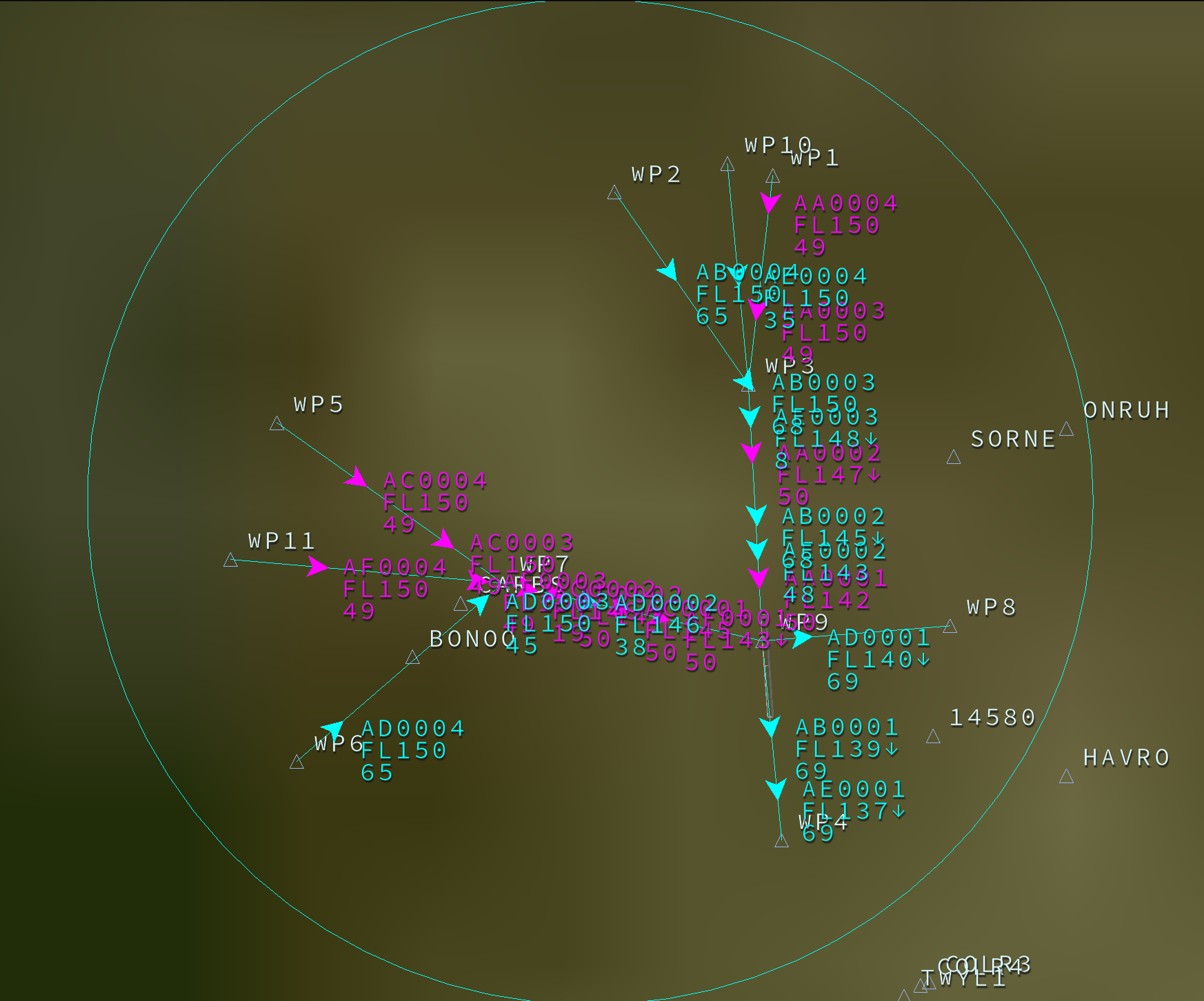}
        \caption{Scenario B}
        \label{fig:scenB}
    \end{subfigure}\hfill
    \begin{subfigure}[t]{0.32\textwidth}
        \centering
        \includegraphics[width=\linewidth]{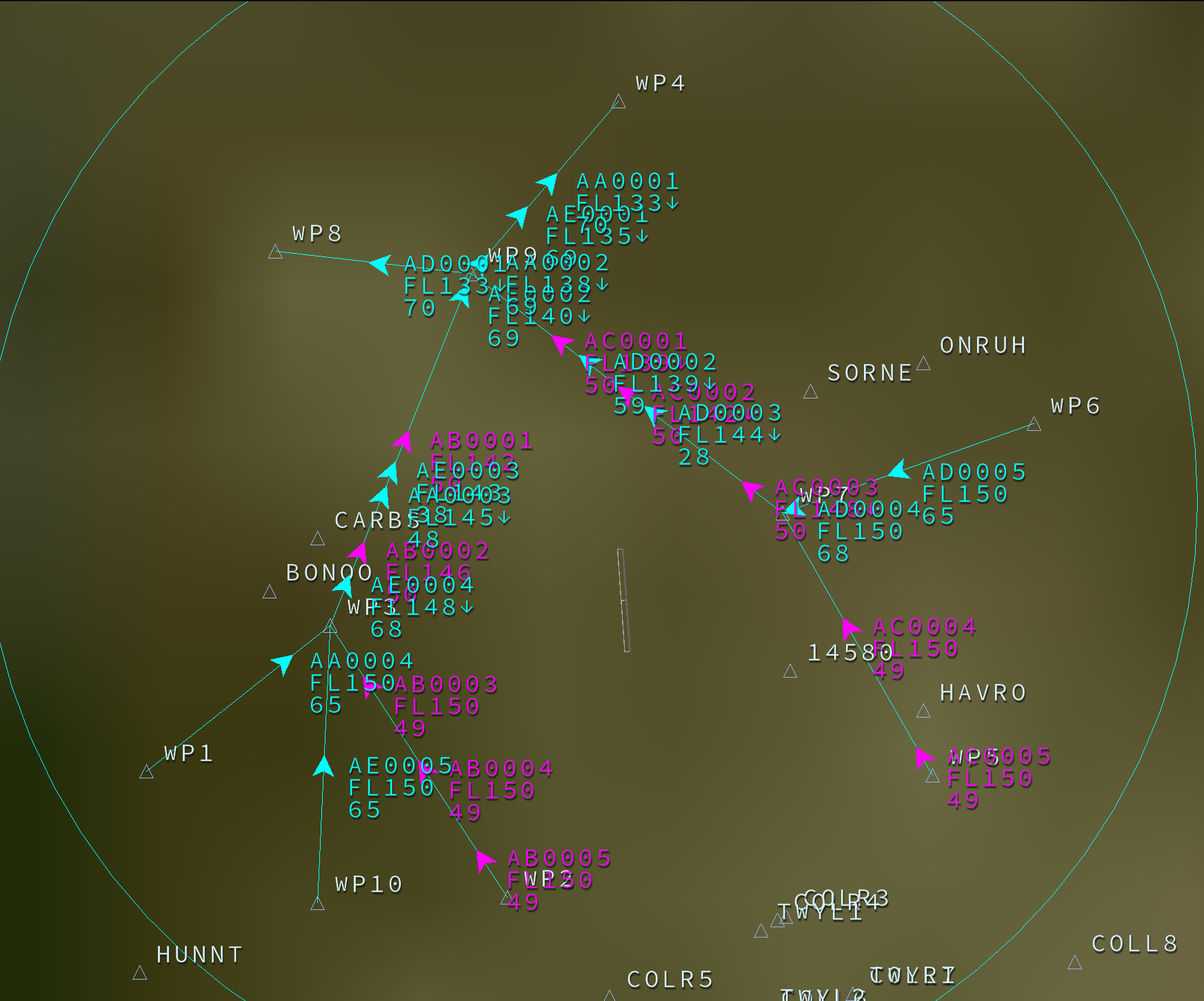}
        \caption{Scenario C}
        \label{fig:scenC}
    \end{subfigure}
    \caption{Traffic snapshots for the three scenarios (A, B, C) used in Table~\ref{tab:bluesky-eval}. The LLM agents and the Rule-based agents are colored in pink and green, respectively. Each scenario has 5-6 routes, each of which hosts 5 agents with random spawning times. Throughout all scenarios, we considered 10 LLM agents, and the rest are Rule-based agents.}
    \label{fig:scenarios_ABC}
\end{figure*}
\newcommand{\best}[1]{\textbf{#1}}
\newcommand{\worst}[1]{#1$^{\dagger}$}

%% file: sec/5_LLM_selection_finetuning.tex
\section{LLM Selection and Fine-Tuning}
\label{sec:llm_selection_finetuning}

For this study, we selected Qwen-Math-7B~\cite{qwen2024qwen2.5, qwen2024math, bai2023qwen} as the pretrained backbone for all fine-tuning experiments. Qwen-Math-7B is a member of the Qwen-2.5 family of transformer-based language models and is optimized for enhanced reasoning, mathematical comprehension, and logical consistency. Unlike general-purpose instruction-tuned models, Qwen-Math-7B incorporates domain-focused pretraining on scientific and quantitative corpora, enabling robust structured reasoning and symbolic manipulation. These characteristics make it well suited for tactical deconfliction tasks, which require reasoning over spatial relationships, safety margins, and action consequences under uncertainty. Throughout this paper, we refer to the pretrained model as the \emph{Base} model.

\textit{\textbf{LoRA Configuration:}} 
As illustrated in Figure~\ref{fig:llm_atc}, we adapt the Base (Frozen) LLM to the tactical deconfliction domain via LoRA-based fine-tuning, implemented using the \texttt{transformers} library with a PyTorch backend. LoRA adapters were applied to the feed-forward projection layers (\texttt{up\_proj}, \texttt{down\_proj}, and \texttt{gate\_proj}) as well as attention projection layers (\texttt{q\_proj}, \texttt{k\_proj}, and \texttt{v\_proj}) to enhance contextual reasoning. For both SFT and GRPO, the  LoRA rank, scaling factor, and dropout were set to 8, 32, 0.05, respectively. The learning rates for SFT and GRPO are set to $10^{-4}$ and $5\times 10^{-6}$, respectively. The rest of the parameters are set to default values in the corresponding Python packages. 
These hyperparameters were chosen to balance adaptation capacity, training stability, and computational efficiency. 

Due to memory limitations, we restricted output generation to 10 tokens to reduce inference time when serving multiple agents. Similarly, GRPO fine-tuning sampled four candidate responses per prompt to compute the advantage function following Eq~(\ref{eq:advantage}) and using maximum temperature to encourage exploration. Both SFT and GRPO training were conducted for a single epoch, requiring approximately 6 and 14 hours, respectively. Optimization was performed using the AdamW optimizer with a cosine learning-rate schedule and warm-up steps to ensure stable convergence. %The SFT stage minimized the negative log-likelihood of human-aligned responses, while the subsequent GRPO stage refined policy behavior through reward-driven optimization.

\textit{\textbf{Reward Function in GRPO:}} 
The reward signal guiding GRPO optimization combines two complementary components: a format reward and an action reward. The format reward, denoted as $r_{\text{format}}$, encourages adherence to the desired response structure by quantifying normalized textual similarity between the generated response $\hat{y}$ and the ground-truth response $y$ via Levenshtein similarity:
\(r_{\text{format}} = \left( 1 - \frac{\Gamma(\hat{y}, y)}{\max(|\hat{y}|, |y|)} \right)^{\gamma},\)
where $\Gamma(\cdot)$ denotes the Levenshtein distance and $\gamma \in [1,\infty)$ controls sensitivity to formatting deviations. The action reward, denoted as $r_{\text{action}}$, enforces decision correctness by verifying whether the action specified in the generated response matches the ground-truth action label: \(r_{\text{action}} = \mathbb{I}\!\left[\text{action}(\hat{y}) = \text{action}(y)\right] - 0.5,\)
where $\mathbb{I}[\cdot]$ is the indicator function. The offset of $-0.5$ centers the reward around zero, penalizing incorrect actions while rewarding correct ones. The overall reward is computed as
\(r(y_k, x) = \lambda_f r_{\text{format}}(y_k, x) + \lambda_a r_{\text{action}}(y_k, x),\)
with weighting coefficients $\lambda_f$ and $\lambda_a$ balancing structural compliance and decision accuracy. 

All experiments were conducted on two NVIDIA RTX~3090 GPUs using mixed-precision training to reduce memory consumption and improve throughput. GRPO training was implemented using the TRL framework. Through this fine-tuning process, Qwen-Math-7B internalizes both rule-based decision logic and context-dependent tactical reasoning, yielding interpretable and safety-aligned decision policies suitable for cooperative multi-agent tactical deconfliction.

\begin{table}[t!]
\centering
\caption{Performance comparison on the evaluation dataset. All numbers are reported in percent (\%).}
\label{tab:evaluation-dataset}
\begin{tabular}{lcccc}
\toprule
\textbf{Model} & \textbf{Accuracy} & \textbf{Precision} & \textbf{Recall} & \textbf{F1-score} \\
\midrule
\textbf{Base}      & 27  & \textbf{75}  & 20  &  31 \\
\textbf{SFT}   & \textbf{88}  & \textbf{75} & \textbf{66}  &  \textbf{69} \\
\textbf{GRPO}  &  53 & \textbf{75} &  40 &  50 \\
\bottomrule
\end{tabular}
\end{table}

%% file: sec/6_Experiments.tex
\section{Experimental Results and Discussions}
\label{sec:results}
Figure~\ref{fig:training-curves} illustrates the training dynamics of the two fine-tuning approaches. The SFT loss curve exhibits stable convergence, indicating effective supervised alignment with human-labeled actions, while the GRPO reward trajectory reflects gradual preference-based policy refinement. These trends suggest that both methods effectively incorporate training signals, albeit through different learning mechanisms. To comprehensively assess the fine-tuned models, we use two strategies: 

\subsection{Evaluation with Datasets}
We first assess the effectiveness of the proposed fine-tuning strategies on a held-out dataset of prompt--response pairs different from the training data. Each sample consists of a natural-language description of a local traffic situation and a corresponding ground-truth tactical action. This evaluation assesses how accurately each model reproduces the desired decision given identical inputs.

During testing, all models were prompted with the same evaluation set, and their generated responses were compared against the reference labels. A prediction was deemed \emph{correct} if the response contained the target action (\textit{Accelerate}, \textit{Hold}, or \textit{Decelerate}); otherwise, it was classified as incorrect. This criterion enables a consistent comparison among the pretrained Base model, the SFT model, and the GRPO fine-tuned model.

Quantitative results on the evaluation dataset are reported in Table~\ref{tab:evaluation-dataset} using standard classification metrics. The Base model achieves an accuracy of 27\%, underscoring the challenge posed by tactical deconfliction for general-purpose LLMs without domain adaptation. In contrast, SFT with LoRA substantially improves performance, achieving an accuracy of 88\% and an F1-score of 69\%, indicating effective alignment with the structured decision patterns encoded in the dataset. The improvement in recall indicates that the SFT model generalizes more reliably across diverse conflict geometries.

The GRPO fine-tuned model attains moderate gains over the Base model, with an accuracy of 53\% and an F1-score of 50\%. While preference-based optimization improves response structure and consistency, its performance remains below that of SFT under the current reward formulation. This outcome suggests that, for this task, direct supervised alignment with human-labeled actions provides a stronger learning signal than relative preference optimization alone.

\begin{table*}[t]
\centering
\caption{Safety and efficiency across configurations and LLM models (mean $\pm$ std, for 10 episodes).
Rates are NMACs/episode. Abbreviations: \textit{L--L} = NMACs between two LLM agents; \textit{L--R} = NMACs between an LLM agent and a Rule-based agent; \textit{All} $=$ \textit{L--L} $+$ \textit{L--R};
\textit{SR} = success rate of LLM agents (fraction completing without NMACs); \textit{Time} = average flight time of successful LLM agents.
\textbf{Bold} indicates best values: lowest NMAC for (All, L--L, L--R), highest SR, and lowest Time among methods with SR $\ge 0.9 \times$ SR$_\text{best}$ for the scenario.}
\label{tab:bluesky-eval}
\scriptsize
\setlength{\tabcolsep}{2.2pt}
\renewcommand{\arraystretch}{1.05}
\begin{tabular}{l ccccc ccccc ccccc}
\toprule
& \multicolumn{5}{c}{\textbf{Base}} & \multicolumn{5}{c}{\textbf{SFT}} & \multicolumn{5}{c}{\textbf{GRPO}} \\
\cmidrule(lr){2-6}\cmidrule(lr){7-11}\cmidrule(lr){12-16}
\textbf{Scen.}
& All & L--L & L--R & SR & Time
& All & L--L & L--R & SR & Time
& All & L--L & L--R & SR & Time \\
\midrule
\textbf{A}
& 3.5$\pm$1.1 & 2.7$\pm$0.5 & 0.8$\pm$1.0 & 0.12$\pm$0.38 & 3.7$\pm$4.1
& \textbf{1.0$\pm$0.8} & \textbf{0.7$\pm$0.7} & \textbf{0.3$\pm$0.5} & \textbf{0.77$\pm$0.29} & \textbf{5.7$\pm$0.6}
& 1.7$\pm$0.5 & 1.1$\pm$0.7 & 0.6$\pm$0.5 & 0.57$\pm$0.31 & 5.2$\pm$0.1 \\
\textbf{B}
& 3.4$\pm$1.1 & 1.8$\pm$1.4 & 1.6$\pm$0.9 & 0.20$\pm$0.56 & 3.3$\pm$7.0
& \textbf{1.9$\pm$1.2} & \textbf{0.9$\pm$0.7} & \textbf{1.0$\pm$0.8} & \textbf{0.62$\pm$0.36} & \textbf{8.1$\pm$0.9}
& 3.0$\pm$0.7 & 1.3$\pm$0.4 & 1.7$\pm$0.5 & 0.27$\pm$0.22 & 6.9$\pm$0.3 \\
\textbf{C}
& 4.0$\pm$0.9 & 2.5$\pm$1.2 & 1.5$\pm$1.2 & 0.05$\pm$0.58 & 1.6$\pm$5.0
& \textbf{1.9$\pm$0.7} & 0.8$\pm$0.6 & \textbf{1.1$\pm$0.9} & \textbf{0.52$\pm$0.37} & \textbf{7.5$\pm$0.7}
& 2.3$\pm$0.8 & \textbf{0.6$\pm$0.5} & 1.7$\pm$0.7 & 0.42$\pm$0.29 & 6.6$\pm$0.1 \\
\bottomrule
\end{tabular}
\end{table*}

\subsection{Evaluation with BlueSky Simulations}

We next evaluate the fine-tuned LLM policies in closed-loop multi-agent simulations using the BlueSky simulator. Figure~\ref{fig:llm_atc} illustrates the \textit{Inference} loop. At each simulation time step, the state information of every LLM-controlled agent is transformed into a structured prompt by the \textit{prompt generator} and passed to the fine-tuned LLM, which outputs the corresponding tactical actions. These actions are then applied to the simulator to update the environment. The process repeats iteratively until all agents exit the scenario. Unlike the dataset-level evaluation, which assesses single-step decision accuracy, this experiment examines emergent system-level behavior, including safety, coordination, and operational efficiency, under realistic multi-agent interactions in unseen scenarios. Table~\ref{tab:bluesky-eval} summarizes safety and performance metrics across three representative traffic scenarios depicted in Figure~\ref{fig:scenarios_ABC}.

Across all scenarios, the pretrained Base model exhibits poor safety and reliability, with high near mid-air collision (NMAC) rates and very low success rates. Success rate is defined as the fraction of LLM agents that complete the scenario without any collision event. These results indicate that zero-shot LLM reasoning, without domain-specific alignment, is insufficient for tactical deconfliction in dense and heterogeneous airspace. In contrast, both fine-tuning strategies substantially improve safety and mission completion, confirming the necessity of domain adaptation for closed-loop deployment.

The SFT model consistently achieves the strongest overall performance across scenarios A, B, and C. It yields the lowest total NMAC rates and the highest success rates, while maintaining reasonable flight times among successful episodes. This behavior suggests that supervised alignment with human-labeled tactical decisions enables the model to internalize safety-oriented heuristics that generalize across diverse conflict geometries. Notably, SFT reduces both LLM--LLM (L--L) and LLM--Rule-based (L--R) NMACs, indicating improved coordination not only among learning agents but also in mixed-policy environments.

The GRPO model demonstrates intermediate performance, consistently improving over the Base model but falling short of SFT in overall safety and reliability. While GRPO reduces NMAC rates and increases success rates relative to the pretrained baseline, its performance varies more strongly across scenarios. In particular, GRPO achieves the lowest L--L NMAC rate in Scenario~C, suggesting that preference-based optimization can enhance coordination among LLM agents in dense traffic. However, this benefit is accompanied by higher L--R NMAC rates and lower success rates compared to SFT, highlighting a trade-off between relative coordination and global safety consistency.

Flight time analysis further illustrates this trade-off. The Base model’s shorter average flight times primarily reflect early episode termination due to NMACs. In contrast, the longer flight times observed for SFT and GRPO correspond to successful mission completion and more conservative deconfliction behavior. Among methods achieving comparable success rates, SFT attains the lowest average flight time, indicating a favorable balance between safety and operational efficiency.

The BlueSky evaluation demonstrates that supervised fine-tuning with human-aligned labels yields the most consistent and reliable closed-loop behavior across heterogeneous scenarios. Preference-based optimization via GRPO offers complementary benefits in specific coordination settings but exhibits reduced robustness under mixed-policy interactions. These results reinforce the importance of human-aligned supervision for deploying LLM-based tactical deconfliction policies in safety-critical airspace \mbox{environments}.

\subsection{Limitations and Broader Impacts}

Despite the encouraging results, several limitations currently constrain the deployment of LLM-based policies in real-time coordinated multi-agent sUAS operations. A primary challenge is inference latency. Even under optimized inference settings, the Base model requires approximately 0.2~seconds to generate a short response for a single agent, with latency scaling linearly with the number of agents and output length. This overhead limits scalability in dense traffic scenarios and restricts the use of more computationally intensive reasoning techniques, such as chain-of-thought prompting or retrieval-augmented generation, which could otherwise enhance decision transparency.

A second limitation concerns prompt sensitivity and stability. LLM behavior is highly dependent on prompt structure, and deviations between the formats used during fine-tuning and inference can lead to degraded performance or partial reversion to pretrained behavior. While structured prompt design mitigates this effect, longer and more descriptive prompts further increase inference time, introducing a trade-off between reasoning richness and real-time responsiveness.

Moreover, reinforcement-based fine-tuning introduces practical constraints. GRPO requires sampling multiple candidate responses per query to estimate relative advantages, resulting in significant computational and memory demands. In this study, hardware limitations constrained the number of sampled responses, likely reducing exploration diversity and training stability. Scaling preference-based optimization for large LLMs therefore remains an open challenge requiring more efficient training strategies and distributed infrastructure.
 
Despite these constraints, this work demonstrates the potential of large language models to support human-aligned and interpretable decision-making in autonomous air traffic coordination, particularly in heterogeneous and dynamic environments. At the same time, the identified limitations underscore the need for careful system-level integration, emphasizing latency-aware design, resource efficiency, and safety assurance. From a broader perspective, the computational cost associated with large-scale fine-tuning motivates continued exploration of lightweight architectures and hybrid symbolic--neural approaches. This study contributes to a growing body of evidence that LLMs can augment, but not yet replace, established decision-making frameworks in real-time, safety-critical applications such as aircraft tactical deconfliction.

%% file: sec/7_conclusion.tex
\section{Conclusion}
\label{sec:conclusion}
This study examined fine-tuned Large Language Models (LLMs) as high-level decision-making policies for tactical deconfliction in dense, heterogeneous, cooperative multi-agent air traffic environments. 

By introducing a ``simulation-to-language'' dataset generation pipeline grounded in interpretable rule-based human decision heuristics, we showed that LLMs, specifically Qwen-Math-7B, can acquire structured, safety-oriented reasoning capabilities for sUAS tactical deconfliction. Using this dataset, we evaluated two complementary parameter-efficient alignment strategies: Supervised Fine-Tuning (SFT) and Group-Relative Policy Optimization (GRPO).

Evaluations on held-out datasets and closed-loop BlueSky simulations demonstrate that SFT provides the most consistent improvements over the baseline LLM in decision accuracy, behavioral stability, and separation safety relative to the pretrained baseline. In contrast, GRPO enables preference-based refinement that improves coordination among LLM agents in certain traffic configurations but exhibits reduced robustness in mixed-policy environments. 
Despite these advances, challenges remain, including inference latency, sensitivity to prompt structure, and the computational demands of reinforcement-style fine-tuning. Overcoming these constraints will be essential for deploying LLM-based tactical deconfliction policies to real-time, large-scale sUAS operations. 

Future work should further benchmark LLM-based approaches against established rule-based and reinforcement-learning-based methods.

%% file: sec/Supplementary_Material.tex
\clearpage
\setcounter{page}{1}

\begingroup
\onecolumn
\maketitlesupplementary

% ============================================
% LISTINGS STYLE FOR CODE/DATA SNIPPETS
% ============================================
\definecolor{codebg}{RGB}{248,248,248}
\definecolor{codeframe}{RGB}{200,200,200}
\definecolor{codecomment}{RGB}{100,100,100}

\lstdefinestyle{stateSnapshot}{
    backgroundcolor=\color{codebg},
    frame=single,
    rulecolor=\color{codeframe},
    basicstyle=\ttfamily\small,
    breaklines=true,
    breakatwhitespace=false,
    columns=fullflexible,
    keepspaces=true,
    showstringspaces=false,
    xleftmargin=0.5em,
    xrightmargin=0.5em,
    aboveskip=1em,
    belowskip=1em,
    captionpos=b,
}
% \title{%
%     \textbf{Fine-Tuning Large Language Models for Autonomous Tactical Deconfliction of Small Unmanned Aerial Systems}\\[0.75em]
%     \Large Supplementary Material
% }

% \author{Anonymous CVPR Submission \\ Paper ID ******}
% \date{}

\noindent This supplementary document accompanies the main paper and provides additional implementation details to support reproducibility. The main paper is fully self-contained; these appendices offer extended technical specifications that complement the methodology described therein.

\vspace{1.5em}

% ============================================
% APPENDIX A: RULE-BASED POLICY
% ============================================
\appendix
\section{Rule-Based Conflict Resolution Policy}
\label{app:rb_policy}

This appendix describes the deterministic rule-based policy used for action selection in the multi-agent flight environment. The policy is hand-crafted and does not involve learning or parameter tuning. At each decision step, an agent selects one of three discrete actions: \emph{Decelerate}, \emph{Hold}, or \emph{Accelerate}. Action selection is governed by the ownship's distance to the next waypoint, the presence and relative position of nearby intruders, route alignment, and speed constraints. All rules are evaluated sequentially and are mutually exclusive after speed constraint enforcement.

\subsection*{Decision Rules}

The decision rules are partitioned into three sets based on the agent's position relative to bottleneck waypoints, as illustrated in Figure~\ref{fig:rules}. The key parameters governing these rules are:
\begin{itemize}[noitemsep]
    \item $d_o^{\mathrm{wp}}$: Distance from the ownship to its next waypoint
    \item $d_o^{\mathrm{safe}}$: Safety distance threshold for triggering deconfliction maneuvers
\end{itemize}

\begin{figure}[h!]
    \centering
    \includegraphics[width=\textwidth, trim=3mm 4mm 3mm 4mm, clip]{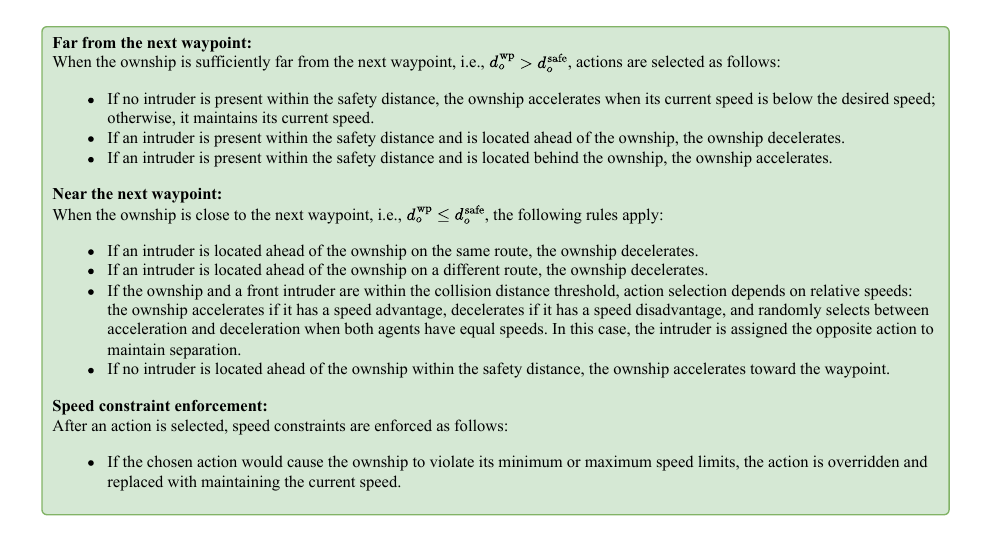}
    \caption{Decision rules for the rule-based policy, organized by ownship proximity to the next waypoint. The policy distinguishes between situations where the ownship is far from the waypoint ($d_o^{\mathrm{wp}} > d_o^{\mathrm{safe}}$) and near the waypoint ($d_o^{\mathrm{wp}} \leq d_o^{\mathrm{safe}}$), with speed constraint enforcement applied as a final override.}
    \label{fig:rules}
\end{figure}

\subsubsection*{Far from the Next Waypoint ($d_o^{\mathrm{wp}} > d_o^{\mathrm{safe}}$)}

When the ownship is sufficiently far from the next waypoint, actions are selected as follows:
\begin{itemize}[noitemsep]
    \item If no intruder is present within the safety distance, the ownship accelerates when its current speed is below the desired speed; otherwise, it maintains its current speed.
    \item If an intruder is present within the safety distance and is located ahead of the ownship, the ownship decelerates.
    \item If an intruder is present within the safety distance and is located behind the ownship, the ownship accelerates.
\end{itemize}

\subsubsection*{Near the Next Waypoint ($d_o^{\mathrm{wp}} \leq d_o^{\mathrm{safe}}$)}

When the ownship is close to the next waypoint, the following rules apply:
\begin{itemize}[noitemsep]
    \item If an intruder is located ahead of the ownship on the same route, the ownship decelerates.
    \item If an intruder is located ahead of the ownship on a different route, the ownship decelerates.
    \item If the ownship and a front intruder are within the collision distance threshold, action selection depends on relative speeds: the ownship accelerates if it has a speed advantage, decelerates if it has a speed disadvantage, and randomly selects between acceleration and deceleration when both agents have equal speeds. In this case, the intruder is assigned the opposite action to maintain separation.
    \item If no intruder is located ahead of the ownship within the safety distance, the ownship accelerates toward the waypoint.
\end{itemize}

\subsubsection*{Speed Constraint Enforcement}

After an action is selected, speed constraints are enforced as a final step:
\begin{itemize}[noitemsep]
    \item If the chosen action would cause the ownship to violate its minimum or maximum speed limits, the action is overridden and replaced with maintaining the current speed (\emph{Hold}).
\end{itemize}

% \newpage

% ============================================
% APPENDIX B: RAW OBSERVATION EXAMPLE
% ============================================
\section{Example Raw Agent Observation}
\label{app:raw_observation}

This appendix presents an example \emph{raw observation record} collected for a single agent at one simulation time step. Listing~\ref{lst:raw_state} shows the exact data structure provided to the rule-based policy prior to action selection, including ownship state variables, information about the two closest front intruders, and the resulting action.

The observation record captures all state information necessary for tactical decision-making, including:
\begin{itemize}[noitemsep]
    \item \textbf{Ownship state:} Position, velocity, heading, route identifier, distance to next waypoint, and speed constraints
    \item \textbf{Intruder information:} Relative positions, velocities, and route identifiers for the two closest front intruders
    \item \textbf{Collision metrics:} Time-to-collision estimates and Euclidean distances to intruders
\end{itemize}

\begin{lstlisting}[style=stateSnapshot,
    caption={Raw observation snapshot for a single agent at one simulation time step. This record is provided to the rule-based policy for action selection and subsequently transformed into a natural-language prompt for LLM training.},
    label={lst:raw_state}]
Ownship info:
  id: A03
  type: Amazon Prime Air - MK30 Model
  lat: 33.137421, lon: -96.861632
  next_wpt_id: WP4
  next_wpt_type: Intersection
  dist_to_nxt_wpt(m): 4759.71
  speed(m/s): 34.98
  min_spd(m/s): 0.0, max_spd(m/s): 41.16
  speed_change_per_second(m/s2): 1.7
  heading(deg): 20.13
  altitude(m): 376.82
  route_id: R_3
  last_action: hold
  num_intruders_ahead: 2
  desired_spd(m/s): 33.44
  time_to_collision_with_intruder1(s): 116.05
  intruder1_on_same_route: True
  did_ownship_have_NMAC: False
  time_to_collision_with_intruder2(s): inf
  intruder2_on_same_route: True
  distance_to_intruder1(m): 1074.77
  distance_to_intruder2(m): 501.82

First closest front intruder info:
  id: D02
  type: Google X-Wing
  lat: 33.14653, lon: -96.85777
  next_wpt_id: WP4
  next_wpt_type: Intersection
  dist_to_nxt_wpt(m): 3685.01
  speed(m/s): 25.72
  min_spd(m/s): 0.0, max_spd(m/s): 30.87
  speed_change_per_second(m/s2): 1.03
  heading(deg): 20.31
  altitude(m): 347.56
  route_id: R_4
  last_action: hold

Second closest front intruder info:
  id: C04
  type: Amazon Prime Air - MK30 Model
  lat: 33.141682, lon: -96.859853
  next_wpt_id: WP4
  next_wpt_type: Intersection
  dist_to_nxt_wpt(m): 4257.95
  speed(m/s): 34.98
  min_spd(m/s): 0.0, max_spd(m/s): 41.16
  speed_change_per_second(m/s2): 1.7
  heading(deg): 20.24
  altitude(m): 355.92
  route_id: R_3
  last_action: hold

Ownship action: Hold.
\end{lstlisting}

\newpage

% ============================================
% APPENDIX C: PROMPT EXAMPLE
% ============================================
\section{Example Prompt for Action Recommendation}
\label{app:prompt_example}

This appendix illustrates the prompt format used for LLM training and inference. The raw observation data (Appendix~\ref{app:raw_observation}) is transformed into a structured natural-language prompt comprising two components:

\begin{enumerate}[noitemsep]
    \item \textbf{System Prompt:} Defines the model's operational role as a tactical deconfliction assistant, specifying the decision context and expected response format.
    \item \textbf{User Prompt:} Describes the current local traffic situation in natural language, including ownship state, intruder information, and relevant spatial relationships.
\end{enumerate}

This translation process converts low-level simulator states into human-readable descriptions that emphasize relative relationships, safety-relevant constraints, and decision context. As a result, the LLM is encouraged to infer tactical reasoning patterns rather than merely learning numerical correlations.

\subsection*{Prompt Structure}

Figure~\ref{fig:prompt-sample} presents a complete example prompt constructed from raw state information. The prompt uses qualitative descriptors (e.g., ``very safe,'' ``very long'') derived from the numerical state values to facilitate natural-language reasoning.

\begin{figure}[h!]
    \centering
    \includegraphics[width=\textwidth, trim=3mm 5mm 3mm 3mm, clip]{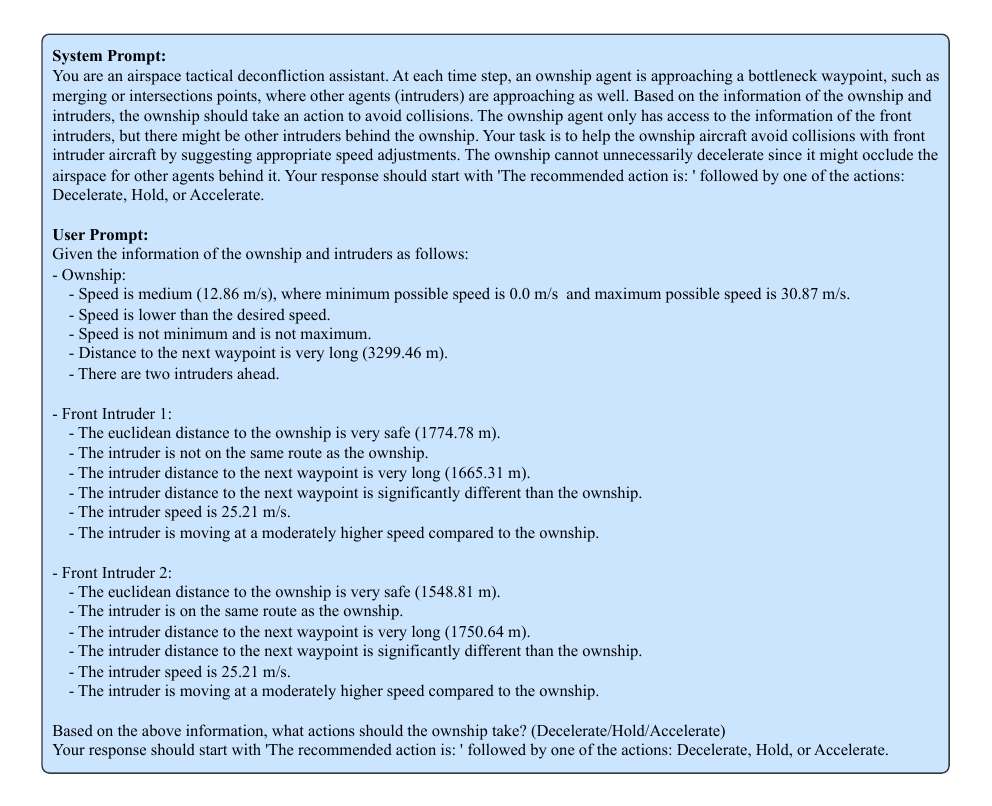}
    \caption{Example prompt for tactical deconfliction at a single time step. The system prompt establishes the model's role and constraints, while the user prompt provides a structured description of the current traffic situation. Qualitative descriptors are derived from numerical thresholds to support natural-language reasoning.}
    \label{fig:prompt-sample}
\end{figure}

\subsection*{Response Format}

The expected response format is shown in Figure~\ref{fig:answer-sample}. The model is trained to produce a brief, structured response beginning with ``The recommended action is:'' followed by one of the three discrete actions: \emph{Accelerate}, \emph{Hold}, or \emph{Decelerate}.

\begin{figure}[h!]
    \centering
    \includegraphics[width=0.5\textwidth, trim=2mm 1mm 2mm 1mm, clip]{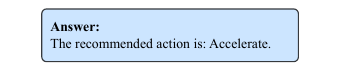}
    \caption{Target response format corresponding to the prompt in Figure~\ref{fig:prompt-sample}. The constrained response format ensures consistent parsing during both training and closed-loop inference.}
    \label{fig:answer-sample}
\end{figure}

\subsection*{Prompt Design Considerations}

Several design choices guide the prompt engineering process:

\begin{itemize}
    \item \textbf{Qualitative descriptors:} Numerical values are converted to qualitative categories (e.g., distance ``very safe'' vs.\ ``critical'') to align with human reasoning patterns and reduce sensitivity to exact numerical values.
    
    \item \textbf{Relative comparisons:} Intruder information emphasizes relative quantities (e.g., ``moving at a moderately higher speed compared to the ownship'') rather than absolute values, supporting transferable reasoning across diverse traffic configurations.
    
    \item \textbf{Constrained output format:} The response format is strictly specified in both the system prompt and the closing instruction, ensuring consistent parsing during evaluation and deployment.
    
    \item \textbf{Safety emphasis:} The system prompt explicitly frames the task in terms of collision avoidance and airspace safety, priming the model toward conservative, safety-oriented decisions.
\end{itemize}

The prompt format is kept consistent across training and inference to ensure behavioral stability. This consistency is critical for maintaining alignment between the fine-tuned model's behavior and the human-aligned supervisory signals encoded in the training dataset.

\endgroup
\twocolumn